\documentclass{article}
\usepackage{ijcai20}

\usepackage{times}
\usepackage{soul}
\usepackage{url}
\usepackage[hidelinks]{hyperref}
\usepackage[utf8]{inputenc}
\usepackage[small]{caption}
\usepackage{graphicx}
\usepackage{amsmath}
\usepackage{amsthm}
\usepackage{booktabs}
\usepackage{algorithm}
\usepackage{algorithmic}
\usepackage{subfigure}
\usepackage{multirow}
\usepackage{amssymb}
\usepackage{stmaryrd}
\usepackage{mathabx}
\usepackage{tabularx}

\newcommand{\vct}[1]{\boldsymbol{#1}} 
\newcommand{\mat}[1]{\boldsymbol{#1}} 

\title{STAS: Adaptive Selecting Spatio-Temporal Deep Features for Improving Bias Correction on Precipitation}

\author{
Yiqun Liu$^1$
\and
Shouzhen Chen$^1$\and
Lei Chen$^2$\and
Hai Chu$^2$\and
Xiaoyang Xu$^1$\and
Junping Zhang$^1$\And
Leiming Ma$^2$
\affiliations
$^1$Shanghai Key Laboratory of Intelligent Information Processing, School of Computer Science, Fudan University\\
$^2$Shanghai Central Meteorological Observation
\emails
\{jpzhang, yqliu17\}@fudan.edu.cn,
malm@typhoon.org.cn,
}

\begin{document}

\maketitle

\begin{abstract}
Numerical Weather Prediction (NWP) can reduce human suffering by predicting disastrous precipitation in time. A commonly-used NWP in the world is the European Centre for medium-range weather forecasts (EC). However, it is necessary to correct EC forecast through Bias Correcting on Precipitation (BCoP) since we still have not fully understood the mechanism of precipitation, making EC often have some biases. The existing BCoPs suffers from limited prior data and the fixed Spatio-Temporal (ST) scale. We thus propose an end-to-end deep-learning BCoP model named Spatio-Temporal feature Auto-Selective (STAS) model to select optimal ST regularity from EC via the ST Feature-selective Mechanisms (SFM/TFM). Given different input features, these two mechanisms can automatically adjust the spatial and temporal scales for correcting. Experiments on an EC public dataset indicate that compared with 8 published BCoP methods, STAS shows state-of-the-art performance on several criteria of BCoP, named threat scores (TS). Further, ablation studies justify that the SFM/TFM indeed work well in boosting the performance of BCoP, especially on the heavy precipitation.
\end{abstract}

\section{Introduction}

Weather forecast plays a crucial role in disaster monitoring and emergency disposal. Numerical Weather Prediction (NWP) based on the equations set of kinetic and thermodynamics~\cite{liu2017kinetic} is often used to cope with sudden climate change and extreme weather beforehand. One of the representatives of progressive NWP in the global is the European Centre for medium-range weather forecasts (EC)~\cite{ran2018evaluation}. However, the predictions of precipitation from EC suffer from some uncertainty intrinsic mechanisms of rainfall, e.g., the elusive physical process of rainfall. Therefore, Bias Correcting on Precipitation (BCoP) is the need to improve the forecast level of EC around a local area. 
\begin{figure}[htbp]
\centering
\subfigure[$t-2$]{
\begin{minipage}[t]{0.3\linewidth}
\centering
\includegraphics[width=1\columnwidth,height=1.2\textwidth]{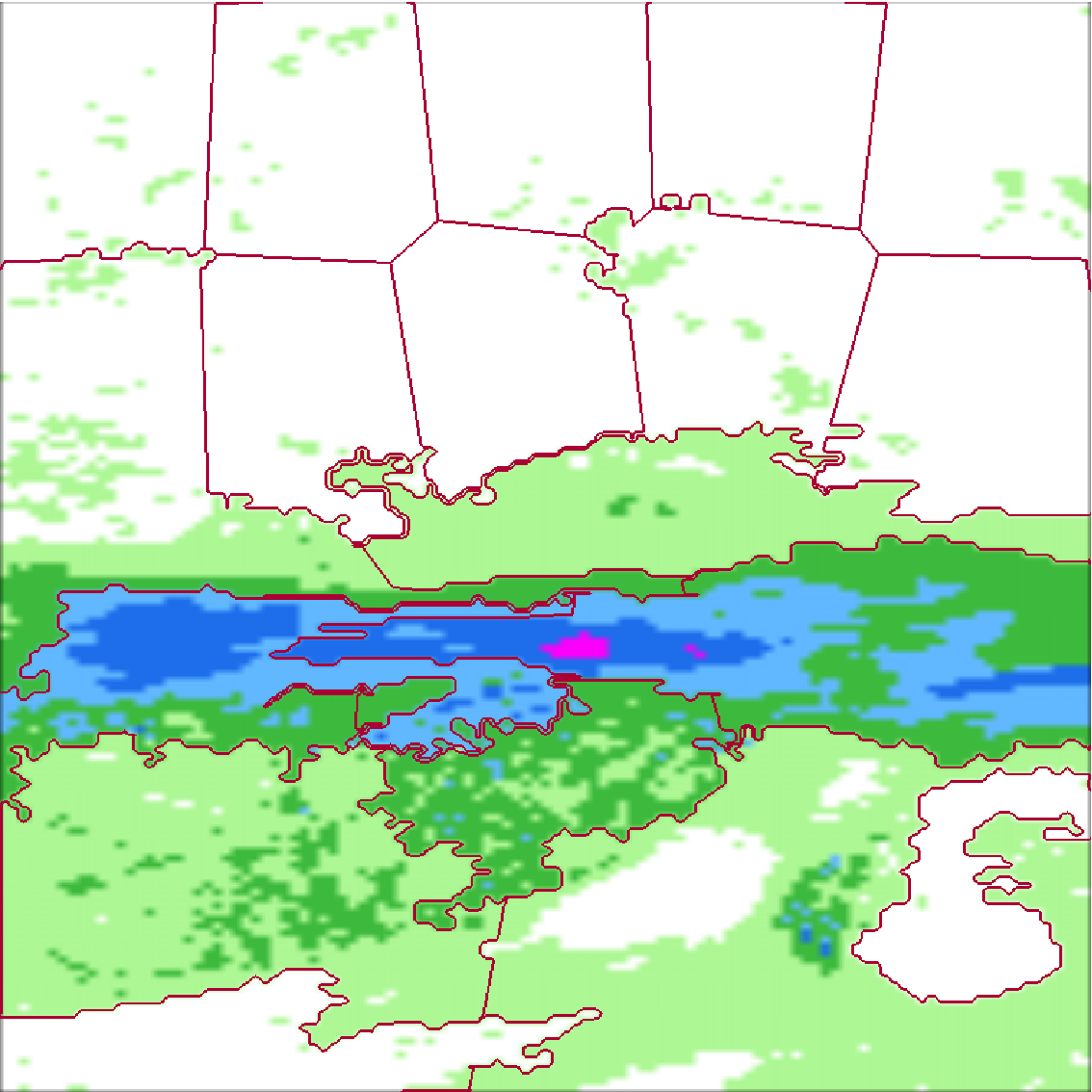}

\end{minipage}%
}%
\subfigure[$t-1$]{
\begin{minipage}[t]{0.3\linewidth}
\centering
\includegraphics[width=1\columnwidth,height=1.2\textwidth]{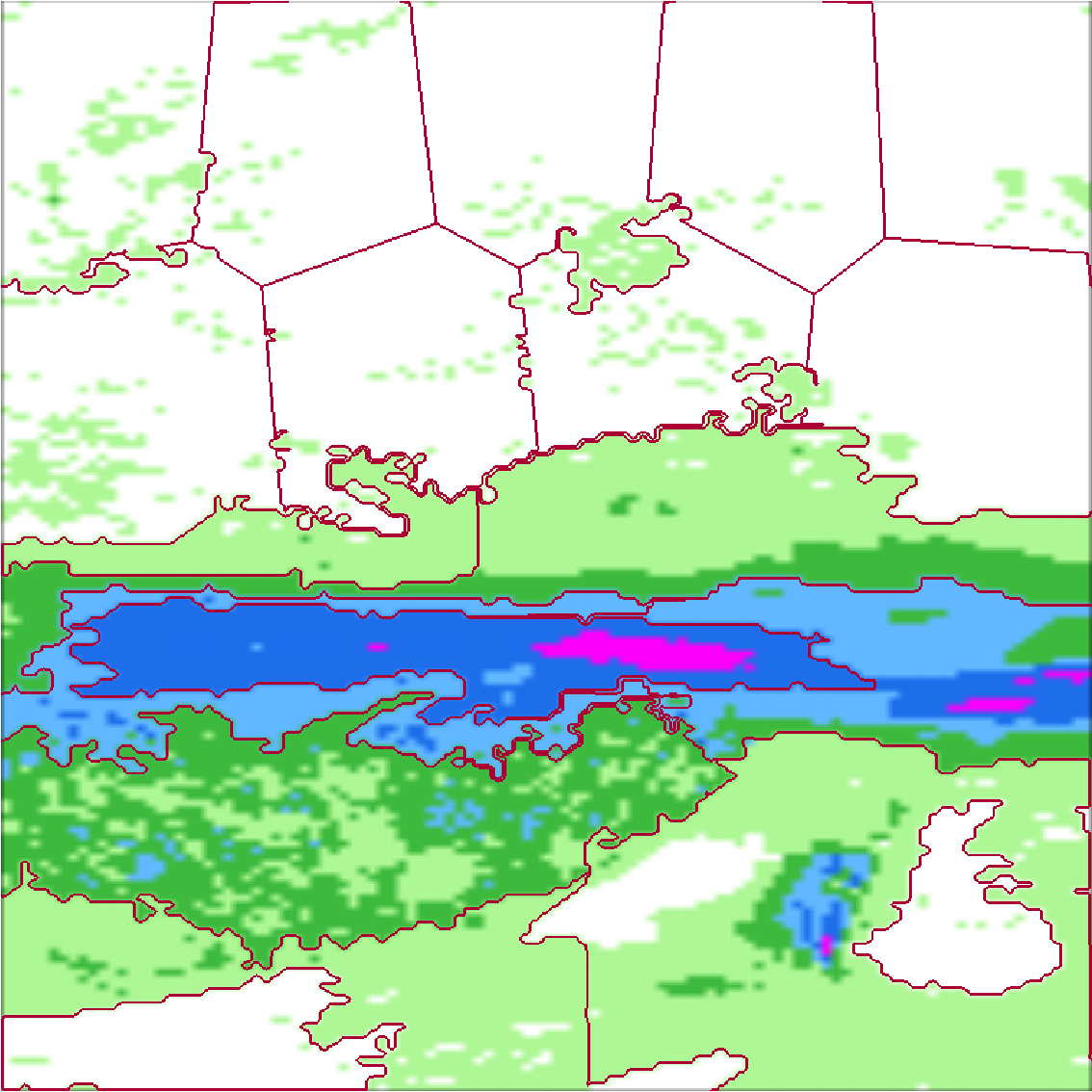}
\end{minipage}%
}%
\subfigure[$t$]{
\begin{minipage}[t]{0.3\linewidth}
\centering
\includegraphics[width=1\columnwidth,height=1.2\textwidth]{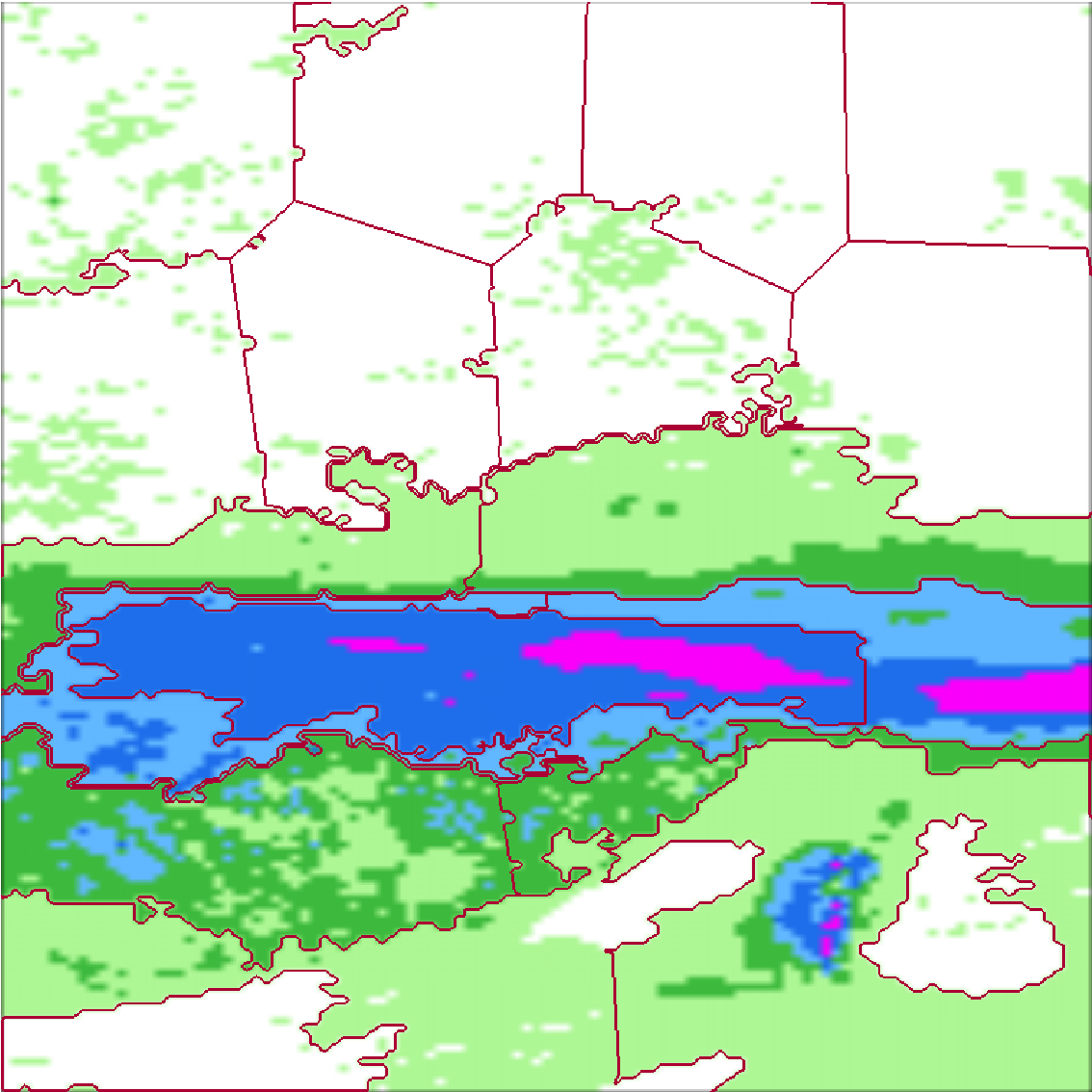}
\end{minipage}
}%
\centering
\caption{The visualization of EC precipitation of Eastern China in the 3 continuous timestamps with region segmentation, which can embody the spatiality and temporal granularity of precipitation.}
\label{ST}
\end{figure}

Classical BCoP can be roughly divided into two categories: regression and parameter estimation. Regression methods can be regarded as a probabilistic model obeying a credible distribution judged by historical prior information such as expert experience~\cite{hamill2008probabilistic}. And the parameter estimating methods figure out a set of optimal parameters in some functions such as Kalman filter~\cite{monache2008kalman} via historical observation for better correcting. Note that in the era of restricted prior information or history observations, these two correcting methods less utilize the spatio-temporal weather regularity in EC~\cite{hamill2012verification}.    

A feasible way to refine the performance of BCoP is to continuously learn the weather dynamic features from EC data and boosting the correcting ability with machine learning algorithms~\cite{srivastava2015wrf}.

Nevertheless, shallow or low-level dynamic representation is not enough to significantly improve the performance of BCoP. Therefore, it is necessary to capture the high-level representation such as spatial and temporal-dependencies from EC data. To clarify this point, we visualize the colored EC precipitation region in the 3 continuous timestamps shown in Fig.~\ref{ST}, and employ a Simple Linear Iterative Clustering (SLIC) is employed for segmenting different precipitation subregions by clustering the pixels in similar semantic information~\cite{achanta2012slic}. From Fig.~\ref{ST}, we can observe that pixels with the same color are naturally segmented into the same region, separated by the purple line, which reflects the spatiality of precipitation. Besides, an obvious movement for the positions of precipitation regions over time reflects varied temporal granularity. 

We thus propose a deep ST Feature Auto-Selective (STAS) Model for learning ST representation. Further, we add two pre-trained modules termed Spatial Feature-selective Mechanism (SFM) and Temporal Feature-selective Mechanism (TFM) to STAS, in which five observations of meteorological elements (precipitation/temperature/pressure/wind/dew) are used to guide the selection of optimal ST scales shown in Fig.~\ref{pipeline} for better extracting ST features. Besides, we integrate a binary classifier so that the precipitation prediction can be more precise. The final prediction is thus achieved by multiplying regression and classified results. The contributions of STAS are summarized as follows:

\begin{itemize}
\item {\bf Spatial Adaptivity} The Spatial Feature-selective Mechanism (SFM)  adaptively selects an optimal spatial scale of specific EC data for capturing the richer spatial representation, which can refine the performance of correcting, especially in the heavy rainfall, indicating its practicability in forecasting mesoscale or large-scale precipitation.  
\item {\bf Temporal Adaptivity} The Temporal Feature-selective Mechanism (TFM) is utilized for automatically choosing the optimal time-lagging sequence of time-series features of EC in line with the minimal loss value for acquiring the better temporal representation.      
\item {\bf Effective} Experiments indicate that our model achieved better prediction performance than the other 8 published methods on a BCOP benchmark dataset, especially for correcting the large-scale precipitation.
\end{itemize}

\section{Related Work}

\subsection{Bias Correcting on Precipitation} 
In this section, we will give a brief survey on Bias Correcting on Precipitation (BCoP) and spatio-temporal (ST) pattern selection. Classical BCoP can be grouped into regression and parameter estimating methods. Regression methods~\cite{hamill2008probabilistic} heavily depend on expert experience, which requires manually setting a threshold for the generation of probability, losing their flexibility and adaptivity. Meanwhile, the parameter estimating methods heuristically assess the key parameters set by trial and error in specific models. Nevertheless, traditional BCoP methods suffer from limited available priors and historical observations, and less utilize the clue of possible motion or dynamics existed in European Centre for medium-range weather forecasts (EC) data. To address these issues, some recently proposed models attempt to capture complex climate patterns from the large-scale EC data and optimize the model in line with observations~\cite{srivastava2015wrf}. However, these methods neglect the potential dependency among variables in EC data, especially the ST dependency.      

\subsection{Spatio-Temporal Pattern on EC Precipitation}
Except for qualitative analysis for reflecting the ST dependencies shown in Fig.~\ref{ST}, it is proved that rainfall value in one location correlates with weather indicators such as temperature, pressure, wind and dew~\cite{yapp1982model}. However, we do not know which scale around this location has a strong connection with the precipitation. Moreover, we know that the system on atmospheric dynamics is a  spatio-temporal evolution system, in which physical field changes over time~\cite{mu2003conditional}. Hence, capturing the ST representation would be a feasible way to improve the performance of BCoP. We consider employing an end-to-end deep-learning model based on multi-scale EC feature~\cite{lin2017feature}, and adaptively select the features that have an optimal scale and take advantage of these features for correcting precipitation more accurately. To our knowledge, there is no report on how to adaptively extract the ST features based deep-learning model for BCoP in literature. This is the first time to adaptively extract ST representation end-to-end.

\section{STAS: A Spatio-Temporal Feature Auto-Selective Model}

\begin{figure*}[!thbp] 
\centering
\includegraphics[width=0.8\textwidth,height=0.28\textwidth]{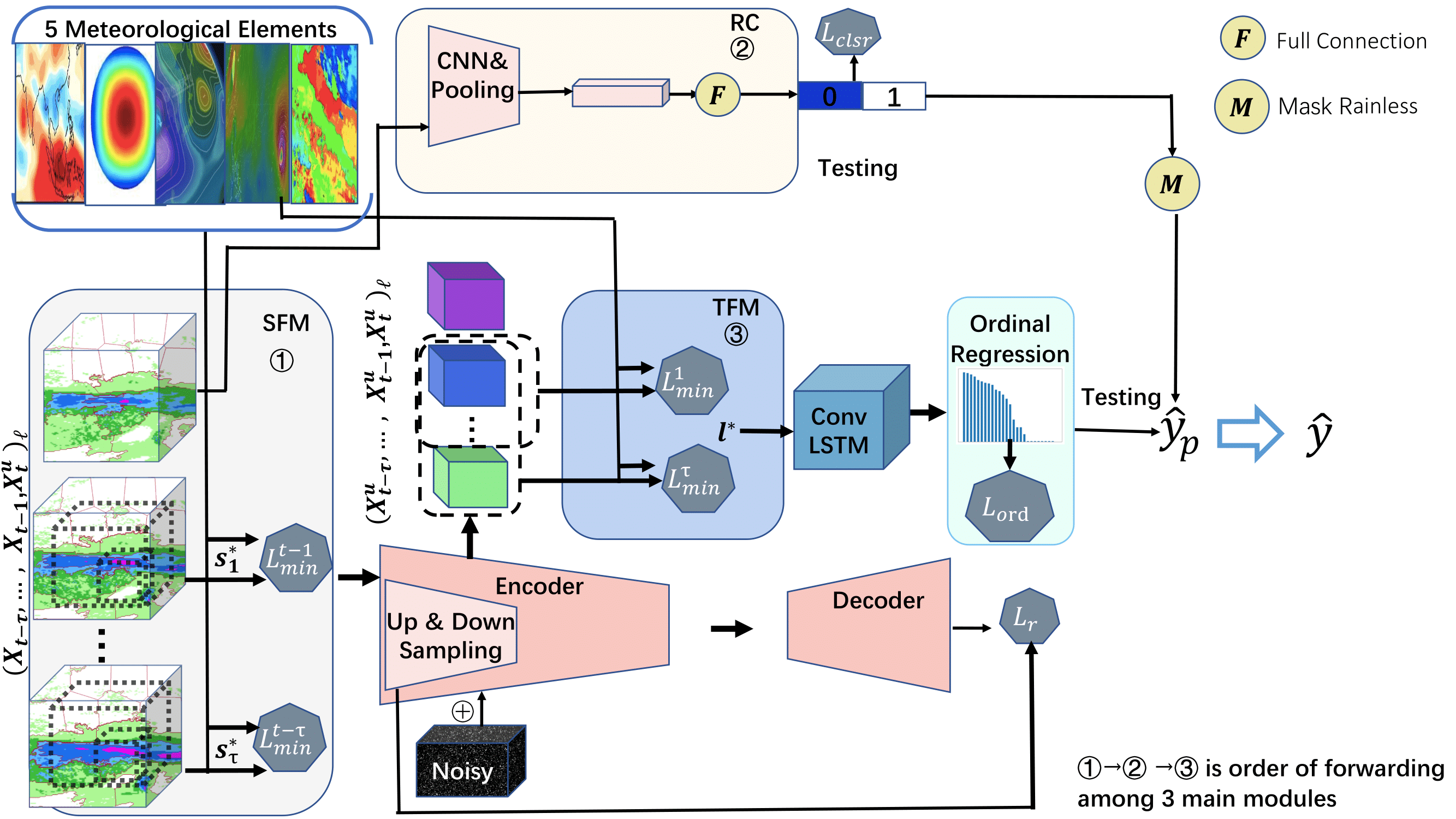}
\caption{The framework is STAS. SFM and TFM is the mechanism of spatial and temporal feature-selective respectively. $s^{*}$ is adaptive spatial size of an EC data in one batch and $\ell^{*}$ is adaptive time-lagging length of the encoded time-series features. $s^{*}$ and $\ell^{*}$ are adaptively adjusted by SFM and TFM respectively. $\mathcal{L}$s are main loss functions in STAS and RC is rainfall classifier. $\mathcal{L}_{min}$s are minimal losses from the process of STAS.}
\label{pipeline}
\end{figure*}

In this section, we will introduce our proposed STAS for automatically selecting the spatial and temporal scales of meteorological features from European Centre for Medium-Range Weather Forecasts (EC) in detail. For better illustration, an overall pipeline of STAS is shown in Fig.~\ref{pipeline}.

\subsection{Notations and Methodology}
First, we define several basic symbols for our method. Assuming that we have a total of $N$ surface observation stations from Eastern China. The EC data around one station can be set as the combination of the refined grid points in the geographical range from \mbox{[$\emph{da} - \omega$, $\emph{do} - \omega$]} to \mbox{[$\emph{da} + \omega$, $\emph{do} + \omega$]}, where \emph{da} and \emph{do} are the latitude and longitude of this station respectively, and $\omega$ is the degree. Furthermore, the time-series EC data from the $i$th station at time level $t$ are defined as $(\mat{X}^{i}_{t})_{\ell} = [\mat{X}_{t}^{u}, \mat{X}_{t-1}, \ldots, \mat{X}_{t-\tau}]_{\ell}$, where $i \in N$, and $\ell$ is the length of time sequence and $\ell = \tau + 1$. $u$ is uniform spatial scale. In this study, the interval of the sequence $\ell$ is 6h. Therefore, $(\mat{X}^{i}_{t})_{\ell}$ can be regarded as a four-dimensional tensor-form input including the dimensions of the channels of features, the length of the sequence, the height and width of features.

With these notations, we roughly build 3 subdivisions for performing Bias Correcting on Precipitation (BCoP) in order as follows:
\begin{eqnarray}\label{eq_pipeline}
 \hat{y}_{tp} &= & OR \left \{LSTM\left ( E\left ([\mat{X}^{u}_{t}, \ldots, \mat{X}^{s^{*}_{\tau}}_{t-\tau}]\right)_{\ell^{*}}\right)  \right\}  \\
 \hat{y}_{rc} &= & RC(\mat{X}^{u}_{t})  \\
 \hat{y}_{t} &= & \hat{y}_{tp} \otimes \hat{y}_{rc}
\end{eqnarray}
where $s^{*}_{\tau}$ denotes the optimized size (height $\times$ width) of the EC data $\mat{X}_{t-\tau}$ through spatial feature-selective mechanism (SFM), which will be introduced in Sec.3.2. Parameter $\ell^{*}$ denotes the refined time-lagging length for time-series features $[(\mat{X}^{i}_{t})_{\ell}]^{u}$ that are encoded (Sec.3.3) to a uniform ($u$) size via temporal feature-selective mechanism (TFM), which will be discussed in Sec.3.4. $E(\cdot)$ is an encoder backbone and $LSTM(\cdot)$ is a stacked ConvLSTM~\cite{xingjian2015convolutional}. The acronym $OR(\cdot)$ means an ordinal regression model~\cite{zhu2018facial} is used for regressing corrected precipitation value in the end. Besides, we utilize the precipitation binary classifier $RC(\cdot)$ for classifying raining or rainless samples. Finally, predicted result $\hat{y}_{t}$ is obtained by multiplying the predicted precipitation value $\hat{y}_{tp}$ and the classified result $\hat{y}_{rc}$.

\subsection{Spatial Feature-Selective Mechanism}
\begin{figure}[t]
\centering
\includegraphics[width=1.4\columnwidth, height=0.28\textwidth]{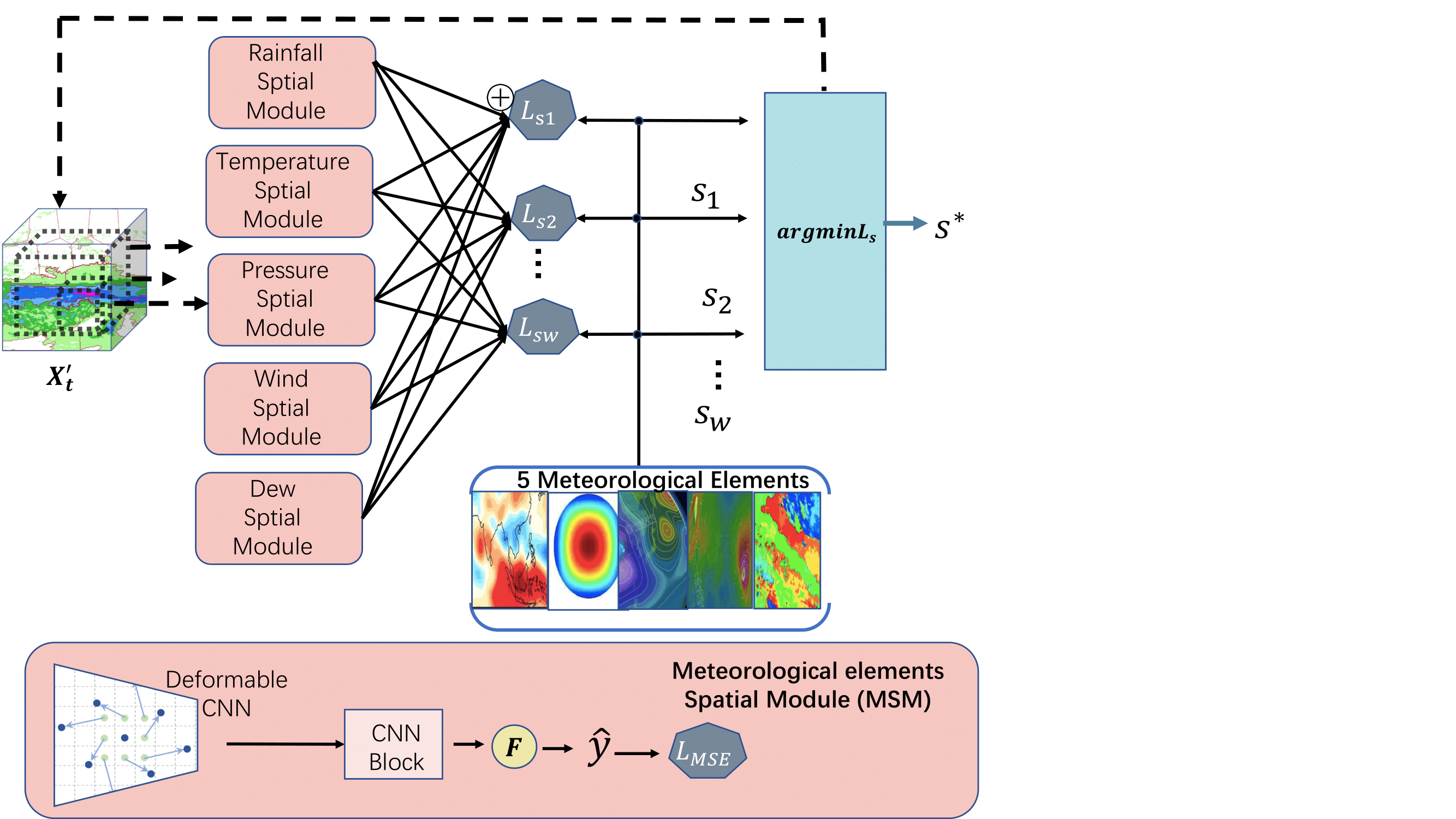} 
\caption{The structure of Spatial Feature-selective Mechanism (SFM), the visualizations of 5 Meteorological Elements (ME) are temperature, pressure, wind, dew and precipitation. $\{ \mathcal{L}_{s1}, \ldots, \mathcal{L}_{sw} \}$ are MSE spatial total losses that are the sum ($\oplus$) of 5 spatial MSE losses from ME spatial modules in different scales. SFM can select the adaptive spatial scale $s^{*}$ of specific EC data in $\mat{X'}_{t}$. The deformable CNN is utilized for helping filters to operate the given pixels that can capture the better representation.}
\label{frameSm}
\end{figure}

\vspace{-0.1em}
When lacking an instructive spatial scale, the EC features centered on all stations are empirically set to be a fixed spatial scale for prediction. However, these rules-of-thumb may impair predictive accuracy when there is a strong connection between features scale and precipitation intensity~\cite{mu2003conditional}. Therefore, we propose a Spatial Feature-selective Mechanism (SFM) to adaptively search the optimal spatial scales of specific EC data based on observations of 5 Meteorological Elements (MEs) including precipitation, temperature, pressure, wind, dew. Concretely, we can find the optimal spatial scales by minimizing the total spatial losses in different scales, which are the summation of 5 spatial MSE losses between predictive MEs and their observations shown in Fig.~\ref{frameSm}. The selection is formulated as:
\begin{eqnarray}\label{spatialMe}
s^{*}  &= & \mathop{\arg\min}_{s}\mathcal{L}_{s} \\ 
\mathcal{L}_{s}  & = & \sum_{i=0}^{n(MEs)}\mathcal{L}_{MSE}(MSM_i(\mat{X'}^{s}_{t}), y_{t}^{i})
\end{eqnarray}where $\mathcal{L}_{s}$ is the spatial total loss in the scale $s$. $n(MEs)$ is the number of MEs. $MSM_i(\cdot)$ is $i$-th ME Spatial Module and $\mat{X'}^{s}_{t}$ is a given EC data that is scale $s$ in timestamp $t$. $y_{t}^{i}$ are the labels of $i$-th ME in timestamp $t$. Therefore, the goal is to search an optimal scale $s^{*}$ shown in Eq \eqref{spatialMe}. Specifically, a deformable CNN layer is introduced for boosting the rep- resentational ability of 5 modules via learning the offsets of filters that are appropriate for capturing better features~\cite{dai2017deformable}. 

\subsection{Backbone with Denoising}
\vspace{-0.1em}
The EC features contain numerous noises~\cite{xu2019towards} that can produce negative side effects for correcting. To solve this issue, we introduce a denoised Encoder-Decoder (E-D) as backbone shown in Fig.~\ref{pipeline}. In the process of encoder, the Gaussian white noise is added to the encoded features that have uniform size $u$ via upsampling and downsampling. We also introduce a reconstruction loss utilized for calculating difference between encoded features and reconstructed features by decoder to optimize denoised ability of encoder as follows:
\begin{equation} \label{reconsLoss}
\mat{W}_{E}^*, \mat{W}_{D}^* = \mathop{\arg\min}_{\mat{W}_{E},\mat{W}_{D}}{\Vert E([\mat{X}^{u}_{t}, \ldots, \mat{X}^{s^{*}_{\tau}}_{t-\tau}], \mat{\epsilon}) - D(Z_{t}) \Vert}
\end{equation}where $\mat{W}_{E}$ and $\mat{W}_{D}$ are parameter matrices from encoder $E(\cdot)$ and decoder $D(\cdot)$ respectively. $[\mat{X}^{u}_{t}, \ldots, \mat{X}^{s^{*}_{\tau}}_{t-\tau}]$ is time-series features with optimized spatial scale $\vct{s^{*}}$ through SFM. $Z_{t}$ is hidden features by $E(\cdot)$. Moreover, $\mat{\epsilon}$ is Gaussian noise.
\subsection{Temporal Feature-Selective Mechanism}
\begin{figure}[t]
\centering
\includegraphics[width=1.3\columnwidth,height=0.28\textwidth]{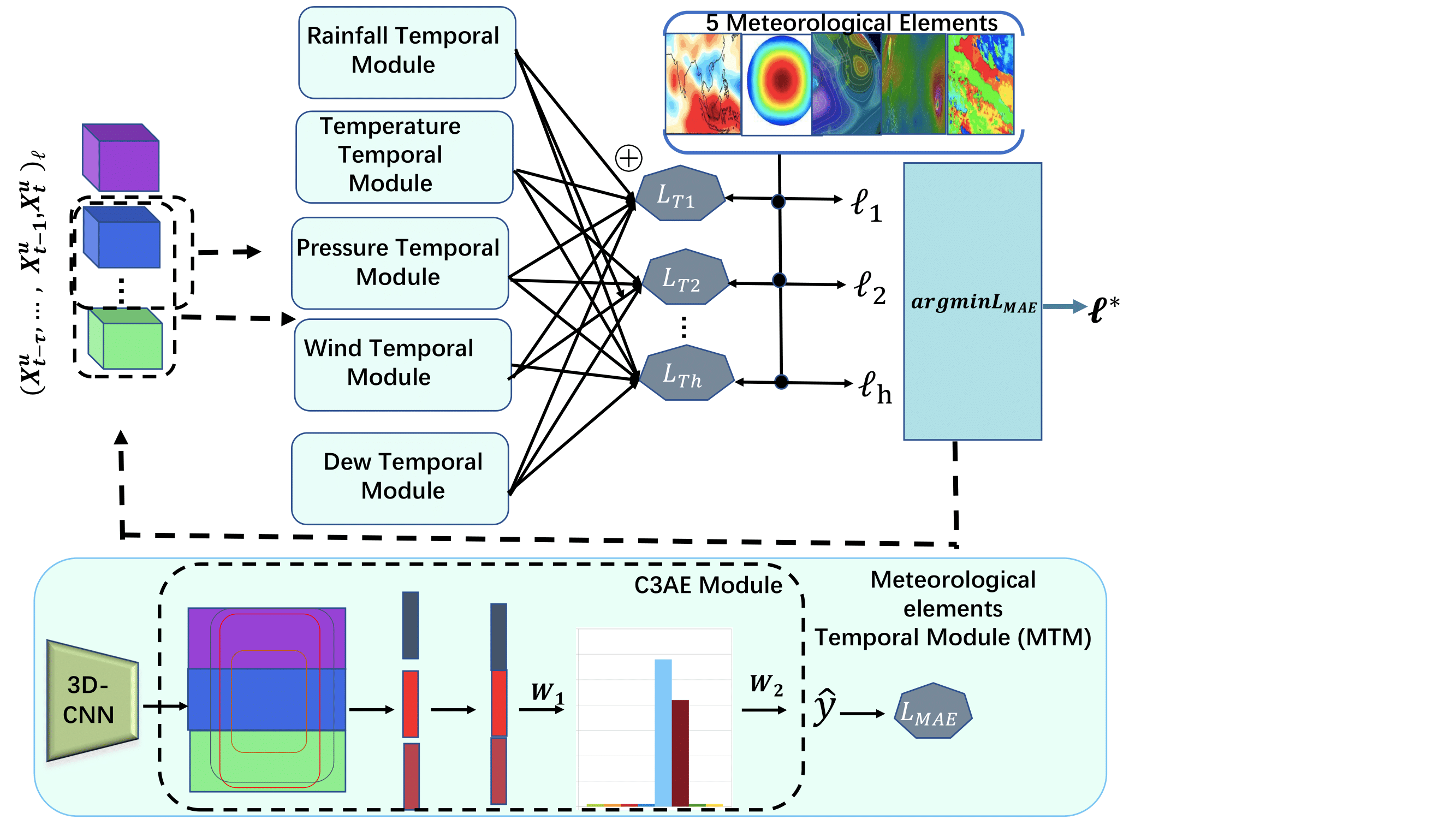} 
\caption{ The structure of Temporal Feature-selective Mechanism (TFM). TFM can select adaptive time-lagging length $\ell^*$ of one time-series features $(\mat{X}^{i}_{t})_{\ell}$. $\{ \mathcal{L}_{T1}, \ldots, \mathcal{L}_{Th} \}$ are MAE temporal total losses that are the sum ($\oplus$) of 5 MAE temporal losses from ME temporal modules in different time-lagging sequence. 3D CNN is utilized for capturing the patial-temporal dependency. Specifically, C3AE is a lightweight rank learning module and suited for regression distribution that has longspan range.}
\label{frameTm}
\end{figure}

\vspace{-0.1em}
It is obvious that rainfall patterns in one station are not only related to EC features around this station, but also closely connected with historical features~\cite{ciach2006analysis}. Therefore, the temporal Feature-selective Mechanism (TFM) is proposed for adaptively acquiring an optimized encoded features $[(\mat{X}^{i}_{t})_{\ell^*}]^{u} = [\mat{X}^{u}_{t}, \ldots, \mat{X}^{u}_{t-\tau}]_{\ell^*}$, , from which more useful temporal representation can be learned. Formally, it is defined as: 
\begin{eqnarray}\label{temporalMe}
l^{*}  &= & \mathop{\arg\min}_{\ell}\mathcal{L}_{T} \\
\mathcal{L}_{T}  & = & \sum_{i=0}^{n(MEs)}\mathcal{L}_{MAE}(MTM_i([(\mat{X}^{i}_{t})_{\ell}]^{u}), y_{t}^{i})
\end{eqnarray}where $\ell^{*}$ is the optimal time-lagging length and $y_{t}^{i}$ are labels of $i$-th MEs in timestamp $t$. $MTM_i(\cdot)$ is $i$-th ME Temporal Module. Same as SFM, TFM can select optimal time-lagging sequence $\ell^{*}$ by finding out minimum temporal total $MAE$ loss, which are the sum of 5 temporal $MAE$ losses between predictions of $MTM(\cdot)$ and labels of MEs in different time-lagging sequences. The detailed structure of C3AE~\cite{zhang2019c3ae} and 3DCNN~\cite{zhang2017learning} are shown in Fig.~\ref{frameTm}. In consideration of maldistribution of precipitation values~\cite{xu2019towards}, an ordinal regression (OR) method~\cite{zhu2018facial} is utilized for outputing regressing value of precipitation $\hat{y}$ shown in Fig.~\ref{pipeline}. OR may solve the problem of longspan range of precipitation values and convert a regression task into a multi-binary classification one to reduce the complexity of regression.

\subsection{Training and Testing}
{\bf Training}\quad In the training phase, we firstly pre-train the Spatial Feature-selective Module (SFM) and Temporal Feature-selective Module (TFM) so that these two modules can predict the MEs accurately. Secondly, we integrate SFM into our framework shown in Fig.~\ref{pipeline} for predicting the MEs from historical EC data $\mat{X'}_{t}^{s}$ and selecting the optimal scale $s^{*}$ that has minimal MSE loss. Similarly, we begin to train Temporal Feature-selective Module (TFM) and Encoder-Decoder (E-D) together, and use a specific ADAM optimizer for BP when adaptive time-lagging length $\ell^{*}$ in one batch is selected by TFM. Meanwhile, the rainy classifier (RC) is cross-trained along with TFM/SFM and E-D.

\noindent{\bf Testing}\quad In the testing phase, SFM plays a role in selecting the optimal scale $s^{*}$ and TFM selects the adaptive time-lagging length $\ell^{*}$, then SFM is forward to calculate probabilities of all classifiers from ordinal regression. We select specific classifiers with large probabilities according to initialized interval $\xi$ and transform these probabilities into regression value of precipitation $\hat{y}_{tp}$ formulated as:
\begin{equation} \label{temporalMe}
\hat{y}_{tp} = \xi * \sum\limits_{v=0}^{c-1}(p_{v}\ge\xi)
\end{equation} where $p_{v}$ is classified probability of the $v$-th binary classifier. Besides, RC is forward to obtain the classified result $\hat{y}_{rc}$. Final predicted result $\hat{y}_{t}$ is required by multiplying SFM result $\hat{y}_{tp}$ and RC result $\hat{y}_{rc}$ as shown in the third equation of Eq \eqref{eq_pipeline} in Sec.3.1.

\section{Experiments}
We conduct all experiments on time-series EC benchmarks collected from 1) the high-resolution version of the public European Centre (EC) dataset and 2) Meteorological Information Comprehensive Analysis and Process System (MICAPS)~\cite{luo2006introduction} that can provide the labels of 5 Meteorological Elements (MEs) including precipitation, temperature, pressure, wind, and dew. Our experiments on Bias Correcting on Precipitation (BCoP) mainly contain two parts. The first part compares the Spatio-Temporal feature Auto-Selective (STAS) model with 8 published machine learning (ML) methods. The second part is a set of ablation experiments on STAS. 

\begin{table}[htp]
\LARGE
\caption{The details of STAS. $N \bigtimes C \bigtimes \ell \bigtimes(h \bigtimes w)_{29\sim3}$ are multi-scale (from 29*29 to 3*3) dimensions of inputs. $(*)_{[.]}$ represent a operator layer and its parameters setting such as filter size from CNN and output scale from Adaptive Pooling (ADP) or Up-Sampling (UpSp). Specifically, the last parameter of Deformable CNN (D-CNN) is spatial dilation rate. Besides, $\oplus Noisy$ is addition operation of Gaussian noise, and $(\cdot)$ is shape of outputs in current module along with pipeline $\rightarrow$.}
\smallskip
\centering
\resizebox{230pt}{25mm}{
\smallskip\begin{tabular}{l|l|l}
\hline
\multicolumn{3}{c}{{\bf Inputs:} $N \bigtimes C \bigtimes \ell \bigtimes(h \bigtimes w)_{29\sim3}$} \\
\hline 
\multirow{2}*{SFM} & $CNN_{[1\times1,0]\rightarrow[3\times3,0]}$ & $(N\bigtimes C\bigtimes (\ell-1)\bigtimes (h\bigtimes w))$ \\
~ & $D-CNN_{[3\times3,1,0.8]\rightarrow[3\times3,1,0.6]} \rightarrow ADPooling_{[1\times1]} \rightarrow FC$ & $\rightarrow (N\bigtimes C \bigtimes(\ell-1)\bigtimes(1\bigtimes1)) \rightarrow(N)\bigtimes(\ell-1) \Rightarrow \vct{s}^*$\\
\hline
\hline
\multirow{2}*{Encoder} & $CNN_{[1\times1,0]\rightarrow[3\times3,0]} \rightarrow ADP_{[16\times16]}$ & $ \rightarrow (N\bigtimes C \bigtimes \ell \bigtimes16\bigtimes16)$ \\
~ & $\oplus Noisy$ & $\oplus \rightarrow (N\bigtimes C \bigtimes \ell \bigtimes18\bigtimes18)$ \\
\hline
\hline
\multirow{2}*{TFM} & $3DCNN_{[3\times3,1]} \rightarrow (ADP_{[1\times1]}+FC) \bigtimes 3$ & $\rightarrow(N\bigtimes C\bigtimes18\bigtimes18) \rightarrow(N\bigtimes6) \bigtimes 3$  \\
~ & $Concat\rightarrow FC\bigtimes2$ & $\rightarrow (N\bigtimes18) \rightarrow (N) \Rightarrow l^{*}$\\
\hline
\hline
Decoder & $UpSp_{[18\times18]}$ & $\rightarrow (N \bigtimes C\bigtimes \ell \bigtimes 18 \bigtimes 18)$ \\
\hline
\hline
RC & $ CNN_{[3\times3,1]}\rightarrow ADP_{[1\times1]} \rightarrow FC$ & $(N\bigtimes C\bigtimes29\bigtimes29) \rightarrow (N\bigtimes C\bigtimes 1\bigtimes1) \rightarrow (N)$ \\
\hline
\multicolumn{3}{c}{{\bf Output:} $N$} \\
\hline
\end{tabular}
}
\label{details}
\end{table}  

\subsection{Datasets and Training Details}
\noindent {\bf EC benchmarks (ECb)} are sliced from a high-resolution version of the public EC dataset~\cite{berrisford2009era}, only covering Eastern China between ranging from June 1st to August 31st for three years (2016-2018). Concretely, ECb consists of 57 weather features (channels) worldwide selected from 601 meteorological factors by Pearson correlation analysis~\cite{benesty2009pearson}. Every feature stems from a grid where each pixel in the grid means a specific location. The spatial and temporal resolutions of ECb are about $111km$ per pixel and 6h per time-level respectively. Then ECb is divided into four datasets for different experiments. The first three datasets are 1) ECb only including Tiny rainfall (ECbT), 2) ECb only including Moderate rainfall (ECbM) and 3) ECb only including Heavy rainfall (ECbH) separated into half-open range of precipitation intensity interval ($[0,1mm)$, $[1-10mm)$, $[10mm,+\infty)$). The last one is ECb Mixed 3 rainfall above (ECbMi) and its mixture ratio of samples is $T:M:H = 9:3:1$.

\noindent {\bf Labels} are observations of 5 Meteorological Elements (MEs) from MICAPS in the specific locations in Eastern China every $6h$.

\noindent {\bf Training details} are shown in Tab.~\ref{details}. The table includes the main structure of layers and initial settings of modules in STAS. Besides, the batch size of training is 256 and the testing is 64. We employ Adam as optimizers for all modules in STAS and the learning rate is $1e-4$. Besides, the weight ratios on 5MEs for spatio-temporal losses are set as $2:1:1:1:1$ in which rainfall is 2 and other elements are 1. The uniform scale after upsampling and downsampling is set as $16*16$. For ordinal regression and C3AE, the ranking intervals for precipitation value are set as 0.5 and 1.5 respectively. Specifically, we set a constant standard $1e-3$ as white Gaussian noise. We test our model in every 6 epochs on training and the max epoch of training is 80. Finally, all experiments are conducted in 8 NVIDIA GPUs.
\begin{table}[htp]
\caption{The 5 criteria between 8 machine learning methods and STAS on ECbMi as in Tab.~\ref{contrastiveI}. $TS_{0.1}$ is TS score in $PI>0.1$. $TS_{1} | TS_{10}$ is $PI>1$ and $PI>10$ respectively. The ECb forecasts are results of predictions from EC benchmarks themselves. SVR is support vector regression, LR is linear regression, MLP is multilayer perceptron, FCN is full convolutional network, FPN is feature pyramid network, LSTM long short-term memory, OBA is ordinal boosting auto-encoder.}
\smallskip
\centering
\resizebox{240pt}{20mm}{
\smallskip\begin{tabular}{c|c|c|c|c|c}
\hline 
       &\multicolumn{5}{|c}{Criteria} \\
\hline 
Methods& MAE & MAPE & $TS_{0.1}$ & $TS_{1}$ & $TS_{10}$ \\
\hline 
ECb forecasts~\cite{ran2018evaluation}  & 1.76 & 17.09 & 0.41 & 0.31 & 0.19 \\
\hline
SVR~\cite{srivastava2015wrf} & 1.67 & 15.81 & 0.48 & 0.37 & 0.1 \\
\hline
LR~\cite{hamill2012verification} & 1.73 & 16.90 & 0.35 & 0.35 & 0.21 \\
\hline
MLP~\cite{yuan2007calibration} & 1.59 & 15.13 & 0.46 & 0.39 & 0.21 \\
\hline
FCN~\cite{xu2019towards} & 1.26 & 12.30 & 0.49 & 0.48 & 0.24 \\
\hline
FPN~\cite{lin2017feature} & 1.15 & 7.38 & 0.56 & 0.51 & 0.27 \\
\hline
LSTM~\cite{xingjian2015convolutional} & 1.21 & 9.8 & 0.52 & 0.48 & 0.24 \\
\hline
OBA~\cite{xu2019towards} & 1.01 & 8.96 & 0.58 & 0.53 & 0.25 \\
\hline
STAS(ours) & {\bf 0.98} & {\bf 5.84} & {\bf 0.75} & {\bf 0.69} & {\bf 0.38} \\
\hline
\end{tabular}
}
\label{contrastiveI}
\end{table} 
\begin{table*}[t]
\centering
\caption{Ablation experiments conducted on ECbMi using $TS_{1}$ and $MAPE$. $\surd$ is defined as an existing component in current ablated STAS for every line of Table. Instead, $\times$ is no this component in the ablation. SFM-MSMs and TFM-MTMs are spatial and temporal meteorological elements modules severally. Besides, D-CNN is deformable CNN and R, T, P, W, and D represents the module of Rainfall, Temperature, Pressure, Wind and Dew respectively.}
\fontsize{8}{9}\selectfont
    \begin{tabularx}{15.5cm}{@{\extracolsep{\fill}}|c|c|ccccc|ccccc|c|c||c|c|}
    \hline
    \multirow{2}{*}{SFM} & \multirow{2}{*}{TFM} & \multicolumn{5}{c|}{SFM-MSMs} & \multicolumn{5}{c|}{TFM-MTMs} & \multirow{2}{*}{D-CNN} & \multirow{2}{*}{C3AE} & \multirow{2}{*}{$TS_1$} & \multirow{2}{*}{$MAPE$} \\

          &  & R & T & P & W & D & R & T & P & W & D &  &  &  &  \\
    \hline
    $\surd$ & $\times$ & $\surd$ & $\surd$ & $\surd$ & $\surd$ & $\surd$ & \multicolumn{5}{c|}{N/A} &$\surd$ & N/A & 0.60 & 8.58 \\
    \hline
    $\times$ & $\surd$ & \multicolumn{5}{c|}{N/A} & $\surd$ & $\surd$ & $\surd$ & $\surd$ & $\surd$ &N/A & $\surd$ & 0.64 & 8.01 \\
    \hline
    $\surd$ & $\surd$ & $\surd$ & $\times$ & $\times$ & $\times$ & $\times$ & $\surd$ & $\surd$ & $\surd$ & $\surd$ & $\surd$ & $\surd$ & $\surd$ & 0.62 & 7.97 \\
    \hline        
    $\surd$ & $\surd$ & $\times$ & $\surd$ & $\surd$ & $\surd$ & $\surd$ & $\surd$ & $\surd$ & $\surd$ & $\surd$ & $\surd$ & $\surd$ & $\surd$ & 0.65 & 8.15 \\
    \hline        
    $\surd$ & $\surd$ & $\surd$ & $\surd$ & $\surd$ & $\surd$ & $\surd$ & $\surd$ & $\times$ & $\times$ & $\times$ & $\times$ & $\surd$ & $\surd$ & 0.65 & 7.03 \\    
    \hline        
    $\surd$ & $\surd$ & $\surd$ & $\surd$ & $\surd$ & $\surd$ & $\surd$ & $\times$ & $\surd$ & $\surd$ & $\surd$ & $\surd$ & $\surd$ & $\surd$ & 0.66 & 7.94 \\   
    \hline        
    $\surd$ & $\surd$ & $\surd$ & $\surd$ & $\surd$ & $\surd$ & $\surd$ & $\surd$ & $\surd$ & $\surd$ & $\surd$ & $\surd$ & $\times$ & $\surd$ & 0.67 & 6.08 \\
    \hline        
    $\surd$ & $\surd$ & $\surd$ & $\surd$ & $\surd$ & $\surd$ & $\surd$ & $\surd$ & $\surd$ & $\surd$ & $\surd$ & $\surd$ & $\surd$ & $\times$ & 0.69 & 6.81 \\    
    \hline        
    $\surd$ & $\surd$ & $\surd$ & $\surd$ & $\surd$ & $\surd$ & $\surd$ & $\surd$ & $\surd$ & $\surd$ & $\surd$ & $\surd$ & $\surd$ & $\surd$ & 0.70 & 5.96 \\ 
     \hline           
    \end{tabularx}
  \label{ablation}
\end{table*}
\subsection{Evaluation Metrics}
MAE and MAPE are regarded as two evaluation criteria for training model.  Here MAE~\cite{willmott2005advantages} is defined as the Mean Absolute Error of corrected precipitation, and MAPE is a variant of MAE without rainless sampless ($<1mm$). In numerical weather prediction, threat score (TS) is a standard criterion for evaluating the accuracy of forecast~\cite{mesinger2008bias} as follows:
\begin{equation} \label{reconsLoss}
TS_{\rho} = H_{\rho}/(H_{\rho} + M_{\rho} +FA_{\rho})
\end{equation}where $H_{.}$ is Hit (correction = 1, truth = 1), $M_{.}$ is Miss (correction = 0, truth = 1), and $FA_{.}$ is False Alarm (correction = 1, truth = 0), in which 1 is rainfall and 0 is rainless. Specifically, $\rho$ is a threshold for splitting the range of Precipitation Intensity (PI) into two intervals and set $[0.1,1,10]$ for three different rainfall cases.
\subsection{Contrastive Experiments on ECbMi}
We list the assessment results of 8 methods and our models on the ECbMi show in Tab.~\ref{contrastiveI}. The reported results are the average of 20 repetitions, each of which is the mean of predicted results on all batches. STAS outperforms all the other methods on five criteria and its $TS_{10}$ is $28.94\%$ higher than the second highest result from OBA in this case. Meanwhile, the performance of FPN and LSTM can extract the spatial and temporal features severally beyond the traditional methods from third line to seventh line. Furthermore, FPN has preferable performance than LSTM. There are mainly two reasons for these phenomena above. 1) the performance of BCoP can be promoted either by learning temporal features or spatial features. 2) As for BCop, the spatial features are more important than temporal features. Furthermore, one possible reason is that FPN can predict rainfall utilized adaptive feature layer that has maximum likelihood~\cite{lin2017feature}, but only learning temporal features in a fixed time scale for LSTM. Besides, the machine learning methods from the fourth to sixth line have somewhat better performance than original ECb forecasts because of utilizing more information from EC data.
\begin{table}[htp]
\caption{The 4 criteria between 3 machine learning methods and STAS on ECbT/ECbM/ECbH divided by precipitation intensity. N/A(Not Applicable) is none of the samples in the current condition.}
\smallskip
\centering
\resizebox{150pt}{20mm}{
\smallskip\begin{tabular}{c|c|c|c|c|c}
\hline 
       &\multicolumn{1}{|c}{Methods} &\multicolumn{4}{|c}{Criteria} \\
\hline 
Ecb & & MAPE & $TS_{0.1}$ & $TS_{1}$ & $TS_{10}$ \\
\hline 
\multirow{4}*{EcbT} & OBA & 3.81 & 0.65 & N/A & N/A \\
~ & LSTM  & 3.79 & 0.55 & N/A & N/A \\
~ & FPN  & 2.14 & 0.60 & N/A & N/A \\
~ & STAS & {\bf 2.01} & 0.78 & N/A & N/A \\
\hline
\multirow{4}*{EcbM} & OBA & 8.39 & 0.53 & 0.49 & N/A \\
~ & LSTM  & 8.01 & 0.51 & 0.47 & N/A \\
~ & FPN  & 6.93 & 0.54 & 0.50 & N/A  \\
~ & STAS & {\bf 4.43} & 0.61 & {\bf 0.59} & N/A  \\
\hline
\multirow{4}*{EcbH} & OBA & 13.44 & 0.21 & 0.21 & 0.09 \\
~ & LSTM  & 12.81 & 0.24 & 0.24 & 0.16 \\
~ & FPN  & 10.79 & 0.28 & 0.28 & 0.20  \\
~ & STAS & {\bf 7.05} & 0.38 & 0.38 & {\bf 0.28}  \\
\hline
\end{tabular}
}
\label{contrastiveII}
\end{table}
\subsection{Contrastive Experiments on ECbT/ECbM/ECbH}
For investigating the influence of ST representation in different precipitation intensity, we compare the performance of the 3 targeted methods and STAS on ECBT, ECBM and ECbH as shown in Tab.~\ref{contrastiveII}.

We prefer to select $MAPE$ instead of $MAE$ because correcting rainfall samples are our main purpose. Overall, The performance of all methods on $TS_{1}$ and $TS_{10}$ decreases, compared with the same $TS$s of these methods in Tab.~\ref{contrastiveI}. The possible reason is that the large rainfall value is hard to correct because of the distribution of longspan and the few limited numbers of samples. Both FPN and LSTM have better performance in $TS_{10}$ than OBA because FPN can capture richer multi-scale spatial features and LSTM can obtain temporal dependency in EC. However, OBA only encodes deep representation. Furthermore, the $TS_{10}$ of LSTM on EcbH is $0.04$ lower than FPN in the same case. One possible explanation is that FPN can automatically select a scale layer with the largest confidence level as the predictive layer in the testing phase, whereas LSTM is a fixed ST scale before testing. Finally, It is worth noting that the $MAPE$s of all methods sharply rise with an increment of precipitation intensity. The reason is that the more samples heavy rainfall has, the bigger contribution MAPE according to its equation. Nevertheless, STAS nearly obtains all SOTA results on 3 Ecbs owing to the learning ability of optimized ST representation.

\subsection{Ablation Experiments}
Here we perform ablation experiments to verify the effectiveness of each new-introduced components in STAS.

\noindent {\bf Impact of SFM and TFM} \quad The first two lines of Tab.~\ref{ablation} show the effectiveness of SFM and TFM. It is obvious that the $TS_{1}$ sharply decreases after removing SFM or TFM, and the $TS_{1}$ of without ($w/o$) SFM is $0.4$ lower than TFM in the same case. One possible reason is that the role of spatial scale is more important in BCoP than that of temporal scale. 

\noindent {\bf SFM vs TFM for ME modules}\quad A similar conclusion can be obtained when we evaluate the SFM/TFM impacts in 5ME modules if we see the results shown in the third line to the sixth line of Tab.~\ref{contrastiveII}. Furthermore, it is obvious that the rainfall (R) module has a greater impact on $MAPE$ than other modules since it utilizes the historical observations of precipitation.     
  
\noindent {\bf Impact of deformable CNN and C3AE}\quad The influences of Deformable CNN (D-CNN) and C3AE are shown in the seventh and eighth lines. MAE of $w/o$ C3AE increases because C3AE is a regression method, quite like ordinal regression for solving the longspan range of precipitation distribution. Besides, the decrement of $TS_{1}$ on $w/o$ D-CNN indicates that D-CNN does work in selecting optimized pixels in the process of learning spatial features.

\begin{figure}[t]
\centering
\includegraphics[width=\columnwidth]{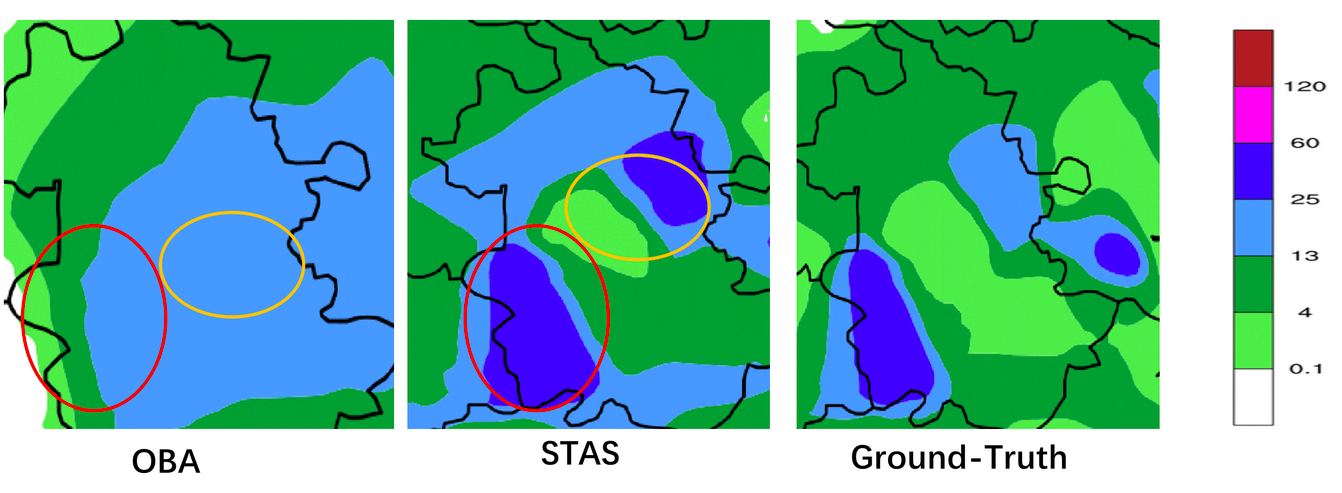} 
\caption{ The visualized comparisons between predictions of precipitation and corresponding ground-truth in the same regions of specific stations. From left to right: the predicted precipitation on OBA and STAS respectively, and observed precipitation. The color-patch bar on the right of Figure is used for distinguishing precipitation intensity by changing light color to high color.}
\label{EC_vs}
\end{figure}   

\subsection{Qualitative Analysis}
We visualize predictive results from 2 methods and corresponding observations in several examples shown in Fig.~\ref{EC_vs}. Among the visualizations, the 2 red ovals on the left and middle pictures show that STAS can almost correct heavy rain (mazarine) accurately while OBA cannot correct exactly. Besides, the 2 orange ovals on the 2 same pictures above reveal that OBA has some errors on calibrating moderate rain (green), but STAS is successful for correcting it. All in all, STAS has better prediction performance than OBA in forecasting mesoscale or large-scale precipitation.

\section{Conclusion}
In this paper, we propose a Spatio-Temporal feature auto-selective (STAS) approach that can automatically extract optimal Spatio-Temporal (ST) features hidden in EC for Bias Correcting on Precipitation (BCoP). Experiments on EC benchmark datasets in Eastern China indicate that STAS achieves the highest threat score (TS) on BCop than other 8 algorithms, and has a strong correcting ability in dealing with different degree of precipitation, especially for heavy precipitation. In the future, we will study how to employ ST mechanisms on more complex precipitation scenarios such as squall line, severe convection and thunderstorm.

\bibliographystyle{named}
\bibliography{ijcai20}

\end{document}